\documentclass[conference]{IEEEtran}
\ifCLASSINFOpdf
  \usepackage[pdftex]{graphicx}
\else
\fi
\usepackage[cmex10]{amsmath}
\usepackage{hhline}
\usepackage{capt-of}
\usepackage{threeparttable}
\usepackage{flushend}

\title{Dropout improves Recurrent Neural Networks for Handwriting Recognition}
\author{
   \IEEEauthorblockN{
      Vu Pham\IEEEauthorrefmark{1}\IEEEauthorrefmark{2},
      Th\'eodore Bluche\IEEEauthorrefmark{1}\IEEEauthorrefmark{3},
      Christopher Kermorvant\IEEEauthorrefmark{1},
      and J\'er\^ome Louradour\IEEEauthorrefmark{1}
      }
   \IEEEauthorblockA{\IEEEauthorrefmark{1} A2iA,  39 rue de la Bienfaisance, 75008 - Paris - France}
   \IEEEauthorblockA{\IEEEauthorrefmark{2} SUTD, 20 Dover Drive, Singapore}
   \IEEEauthorblockA{\IEEEauthorrefmark{3}LIMSI CNRS, Spoken Language Processing Group, Orsay, France}
}
\begin{document}
\maketitle
\begin{abstract}
%By randomly dropping units in deep neural networks during training, dropout is very efficient in preventing those models from overfiting and provides significant improvements in the context of Convolutional networks []. This paper shows that the same approach can be applied for recurrent neural networks which currently hold the best known results in unconstrained handwriting recognition. 

Recurrent neural networks (RNNs) with %enhanced by
Long Short-Term memory cells currently
hold the best known results in unconstrained handwriting recognition.
We show that their performance can be greatly improved using
\emph{dropout} - a recently proposed regularization method for deep architectures.
While previous works showed that dropout gave superior performance in the context of convolutional networks, % \cite{krizhevsky2012ImageNet} JL: avoid citation in abstract
it had never been applied to RNNs.
In our approach, dropout is carefully used in the network
so that it does not affect the recurrent connections,
hence the power of RNNs in modeling sequences is preserved.
Extensive experiments on a broad range of handwritten
databases confirm the effectiveness of dropout on deep architectures
even when the network mainly consists of recurrent and shared connections.

% making it a promising alternative regularization method, in stead of L1 or L2 --> should we compare with L1/L2? too late...

\end{abstract}
\section{Introduction}
\label{sec:intro}

% VPH: Intro RNN-LSTM-CTC in Handwriting recognition
% I have no idea what  to write here, so I give 
% a rough description of handwriting recognition system
% Please correct me if I am wrong.
% JL: That's OK! :-)
Unconstrained offline handwriting recognition is the problem of recognizing
long sequences of text when only an image of the text is available.
The only constraint in such a setting is that the text is written
in a given language. % known by the system.
Usually a pre-processing module is
used to extract image snippets, each contains one single word or line,
which are then fed into the recognizer. A handwriting recognizer, therefore,
is in charge of recognizing one single line of text at a time. Generally,
such a recognizer should be able to detect the correlation between
characters in the sequence, so it has more information about the 
local context and presumably provides better performance. Readers
are referred to \cite{Plamondon2000HandRecogSurvey} for an extensive review
of handwriting recognition systems.

Early works typically use a Hidden Markov Model (HMM) \cite{Marti:2001:USL:505741.505745}
or an HMM-neural network hybrid system \cite{Marukatat01sentencerecognition, Senior1998}
for the recognizer. However, the hidden states of HMMs
follow a first-order Markov chain, hence
they cannot handle long-term dependencies in sequences.
Moreover, at each time step, HMMs can only select one hidden state, hence an HMM
with $n$ hidden states can typically carry only $\log\left(n\right)$ bits
of information about its dynamics \cite{Ghahramani:1997:FHM}.

Recurrent neural networks (RNNs) do not have such limitations and were shown
to be very effective in sequence modeling. With
their recurrent connections, RNNs can, in principle, store representations of
past input events in form of activations,
allowing them to model long sequences with complex structures.
% Nonetheless, RNNs are in essence deep in the temporal (or spatial) domain.
% They can also have multiple layers. %, hence the unrolled RNN is deep at each recurrence step.
RNNs are inherently deep in time and can have many layers, both make training 
parameters a difficult optimization problem.
% This makes optimization of RNNs a particularly hard problem:
The burden of exploding and vanishing gradient was the reason for
the lack of practical applications of RNNs until recently~\cite{Bengio94VanishingGradient,Hochreiter1998RnnDifficulty}.

%Lately, an advance in training RNNs called Long Short-term Memory (LSTM)
%cells was proposed. LSTM is carefully designed memory cells intergrated in RNNs
Lately, an advance in designing RNNs was proposed, namely Long Short-Term Memory (LSTM)
cells. LSTM are carefully designed recurrent neurons
which gave superior performance in a wide range of sequence modeling problems.
In fact, RNNs enhanced by LSTM cells \cite{Graves_Schmidhuber2008}
won several important contests \cite{Menasi2012,Nion2013,A2iASystem_OpenHart2013} 
and currently hold the best known results in handwriting recognition.

% Intro dropout
Meanwhile, in the emerging deep learning movement, dropout was used to effectively
prevent deep neural networks with lots of parameters from overfitting. It is shown
to be effective with deep convolutional networks \cite{Hinton2012Dropout, krizhevsky2012ImageNet, Deng2013DropoutConvNN},
feed-forward networks \cite{Dahl2013LVCSRDropout, Li2013Dropout, Seltzer2013Dropout} but, to the best of our knowledge, has never been applied to RNNs.
Moreover, dropout was typically applied only at fully-connected layers \cite{Hinton2012Dropout, Wan_dropc_2013},
even in convolutional networks \cite{krizhevsky2012ImageNet}.
In this work, we show that dropout can also be used
in RNNs at some certain layers which are not necessarily fully-connected. The choice of
applying dropout is carefully made so that it does not affect the recurrent connections,
therefore without reducing the ability of RNNs to model long sequences.

Due to the impressive performance of dropout, some extensions of this technique were proposed,
including DropConnect~\cite{Wan_dropc_2013},
%a generalized version of dropout which randomly drops weights instead of hidden units,
Maxout networks~\cite{Goodfellow_maxout_2013},
%a new nonlinearity that need dropout to be well-fitted%and provides compelling performance on many classification problems,
and an approximate approach for fast training with dropout \cite{Sidaw13icml}. 
In \cite{Wan_dropc_2013}, a theoretical generalization bound of dropout was also derived.
%The most important result of this bound is that the complexity of the \emph{dropped} network is a linear function of the dropout probability.
In this work, we only consider the original idea of dropout~\cite{Hinton2012Dropout}.

Section \ref{sec:RNN-intro} presents the RNN architecture designed for handwriting recognition.
Dropout is then adapted for this architecture as described in Section \ref{sec:dropout-RNN}. Experimental results
are given and analyzed in Section \ref{sec:Experiments}, while the last section is dedicated for conclusions.

\section{Recurrent neural networks for Handwriting recognition}\label{sec:RNN-intro}

% I have the black-and-white version of this figure (net_bw.pdf)
% and I think the BW version is more professional (especially when shown in a scientific paper)
\begin{figure*}
\centering
\includegraphics[width=0.85\textwidth]{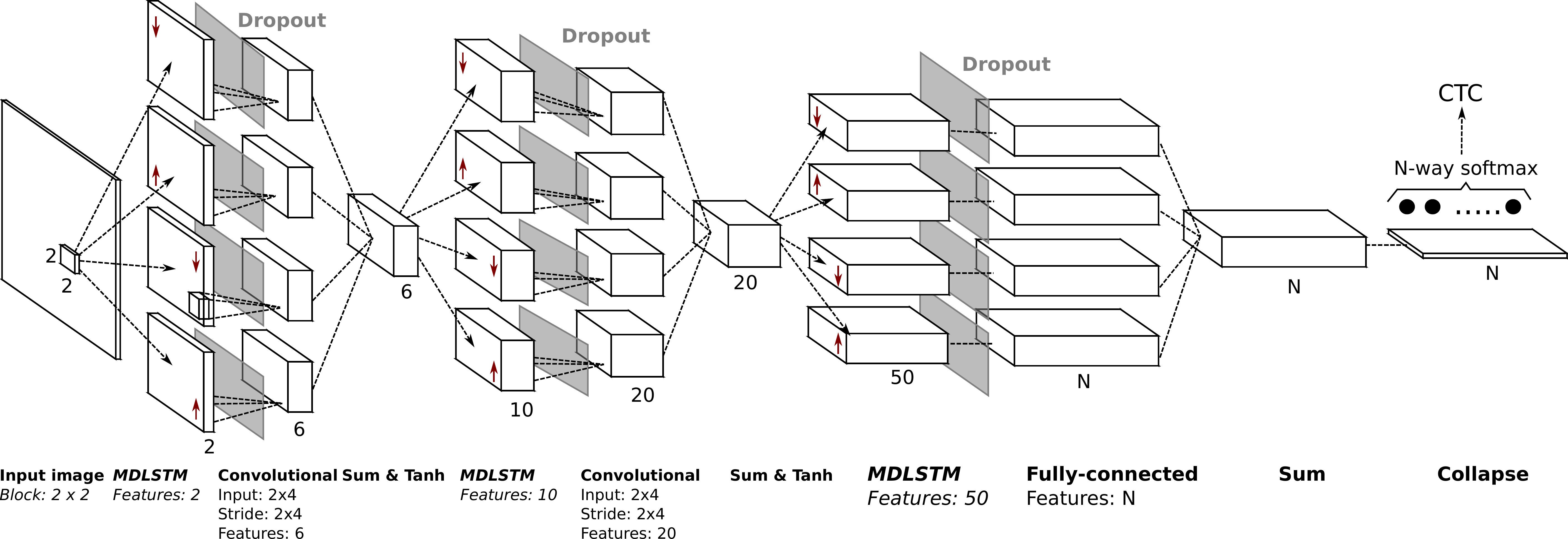}
\caption{The Recurrent Neural Network considered in this paper, with the places where dropout can be applied.}
\label{fig:full-system}
\end{figure*}

The recognition system considered in this work is depicted in Fig.~\ref{fig:full-system}.
The input image is divided into blocks of size $2\times 2$ and fed
into four LSTM layers which scan the input in different directions indicated by corresponding
arrows. The output of each LSTM layer is separately fed into convolutional layers of
6 features with filter size $2\times 4$. This convolutional layer is applied without overlaping
nor biases. % nor pooling.
It can be seen as a subsampling step, with trainable weights rather than
a deterministic subsampling function. % as in other architectures. 
The activations of 4 convolutional layers are then summed element-wise and squashed
by the hyperbolic tangent (tanh) function. This process is repeated twice with
different filter sizes and numbers of features, and the top-most layer
is fully-connected instead of convolutional. The final activations are summed
vertically and fed into the softmax layer. The output of softmax is processed by
Connectionist Temporal Classification (CTC)~\cite{Graves06connectionisttemporal}.

This architecture was proposed in \cite{Graves2008RnnHandwriting},
but we have adapted the filter sizes for input images
at 300 dpi. There are two key components enabling this architecture to give superior performance:
\begin{itemize}
\item \emph{Multidirectional LSTM} layers~\cite{Hochreiter1997LSTM}.
LSTM cells are carefully designed recurrent neurons with multiplicative gates to store 
information over long periods and forget when needed.
Four LSTM layers are applied in parallel, each one with a particular scaning direction. In this way
the network has the possibility to exploit all available context. % against all possible directions
\item \emph{CTC} is an elegant approach for computing the Negative Log-likelihood
for sequences, so the whole architecture is trainable
% where each input image is aligned directly to the output sequence, without any post-processing stage.
without having to explicitly align each input image with the corresponding target sequence.
\end{itemize}

In fact, this architecture was featured in our winning entry of the Arabic handwriting recognition competition 
OpenHaRT 2013~\cite{A2iASystem_OpenHart2013}, where
such a RNN was used as the optical model in the recognition system. % whose outputs was processed by a lexicon and a language model.
In this paper, we further improve the performance of this optical model using dropout
as described in the next section.

%\begin{figure*}[h]
%\begin{minipage}[t]{0.7\textwidth}
%\includegraphics[width=1\textwidth]{figs/net}
%\caption{The deep RNN considered in this paper}
%\label{fig:full-system}
%\end{minipage}
%\hfill{}
%\begin{minipage}[b][1\totalheight][t]{0.3\textwidth}
%\includegraphics[width=1\textwidth]{figs/rnn_unroll_dropout}
%\caption{Dropout is only applied to \emph{feed-forward} connections in RNNs.}
%\label{fig:rnn-dropout}
%\end{minipage}
%\end{figure*}

\section{Dropout for recurrent neural networks} \label{sec:dropout-RNN}

\begin{figure}[h]
\centering
\includegraphics[width=0.85\columnwidth]{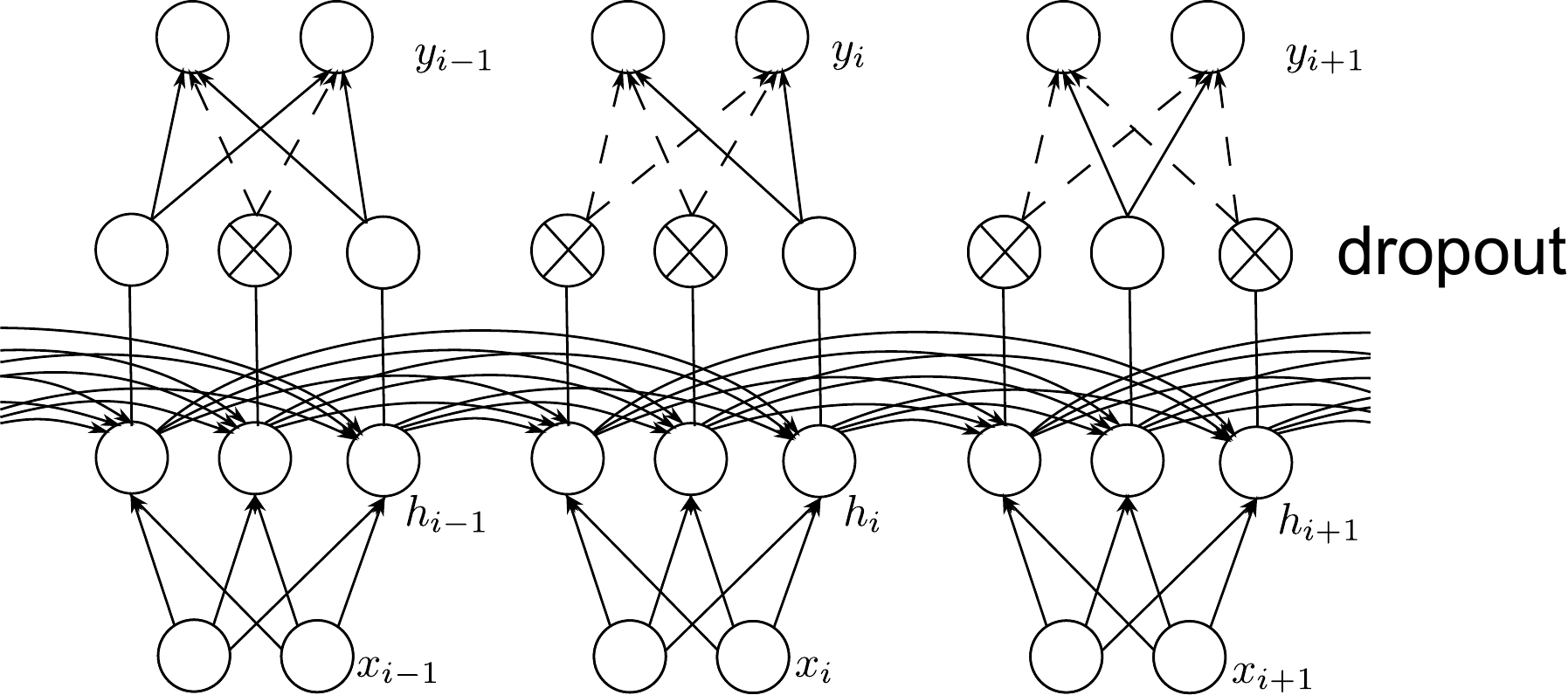}
\caption{Dropout is only applied to \emph{feed-forward} connections in RNNs.
The \emph{recurrent} connections are kept untouched. This depicts
one recurrent layer ($h_i$) with its inputs ($x_i$), and an output layer ($y_i$)
which can comprise full or shared connections.
The network is unrolled in 3 time steps to clearly show the recurrent connections.
}
\label{fig:rnn-dropout}
\end{figure}

% we do not drop recurrent connections etc...
% not sure if we should formally present dropout here...
Originally proposed in \cite{Hinton2012Dropout}, dropout involves randomly removing some hidden units in
a neural network during training but keeping all of them during testing.
More formally, consider a layer with $d$ units and let $\mathbf{h}$ be a $d$-dimensional
vector of their activations. When dropout with probability $p$ is applied at this layer,
some activations in $\mathbf{h}$ are dropped:
$\mathbf{h}^{\text{train}}=\mathbf{m}\odot\mathbf{h}$,
where $\odot$ is the element-wise product, and $\mathbf{m}$ is a binary mask
vector of size $d$ with each element drawn independently from $m_{j}\sim\text{Bernoulli}\left(p\right)$.
During testing, all units are retained but their activations are weighted by $p$:
$\mathbf{h}^\text{test}=p\mathbf{h}$.
Dropout involves a hyper-parameter~$p$, for which a common value is~$p=0.5$.

% In RNNs, %especially the architecture presented in Section \ref{sec:RNN-intro},
% the recurrent connections %in LSTM layers
% play an important role. %in propagating its activations through recurrence steps.
We believe that
random dropout should not affect the recurrent connections in order to conserve the
ability of RNNs to model sequences. This idea is illustrated in Fig.~\ref{fig:rnn-dropout},
where dropout is applied only to \emph{feed-forward} connections and not to
\emph{recurrent} connections. With this construction, dropout can be seen as a
way to combine high-level features learned by recurrent layers. 
Practically, we implemeted dropout as a separated layer whose output is identical to
its input, except at dropped locations $\left(m_j = 0\right)$. With this implementation,
dropout can be used at any stage in a deep architecture, providing more
flexibility in designing the network.

Another appealing method similar to dropout is \emph{DropConnect}~\cite{Wan_dropc_2013},
which drops the connections, instead of the hidden units values. %, which is a generalization of dropout.
However DropConnect was designed for fully-connected layers,
where it makes sense to drop the entries of the weight matrix.
In convolutional layers, however, the weights are shared, so there are only a few
actual weights. If DropConnect is applied at a convolutional layer with $k$ weights, 
it can sample at most $2^k$ different models during training.
In contrast, our approach drops the input of convolutional layers.
Since the number of inputs is typically much greater than the number
of weights in convolutional layers, dropout in our approach samples from a
bigger pool of models, and presumably gives superior performance.
%If the layer after dropout is fully-connected, our approach is similar to DropConnect%
%\footnote{Technically speaking, in fully-connected layers, our approach can only sample from $2^n$ models, where $n$ is the number of inputs, when DropConnect can sample from at most $2^{(n^2)}$ models.}%.
%However our work extends the idea of dropout to make it work with recurrent layers, especially when many kinds of layers are involved in the architecture.

In \cite{Mesnil13RNNDropout}, dropout is used to regularize a bi-directional RNN, but the network
has only one hidden layer, there are no LSTM cells involved,
and there is no detail on how to apply dropout to the RNN.
%
%% JL: did not understand the point here
In \cite{Deng2013DropoutConvNN}, dropout is used in a convolutional neural network
but with a smaller dropout rate because the typical value $p=0.5$ might slow down the
convergence and lead to higher error rate.
In this paper, our architecture has both covolutional layers and recurrent layers.
The network is significantly deep, and we still find the typical dropout rate
$p=0.5$ yielding superior performance. This improvement can be
attributed to the way we keep recurrent connections untouched when applying dropout.

Note that previous works about dropout seem to favor rectified linear units (ReLU)~\cite{krizhevsky2012ImageNet}
over \emph{tanh} or \emph{sigmoid} for the network nonlinearity since it provides
better covergence rate. % in shorter training time.
In our experiments, however,
we find out that ReLU can not give good performance in LSTM cells, hence we keep
\emph{tanh} for the LSTM cells and \emph{sigmoid} for the gates.

% TODO: JL might wanna elaborate on dropout and nonlinearities in RNN?

% DEFINITION OF NAMES IN THE TABLES
\def\CER{\textbf{CER}}
\def\WER{\textbf{WER}} % (\%)
\def\Rimes{\textbf{Rimes}}
\def\IAM{\textbf{IAM}}
\def\OpenHaRT{\textbf{OpenHaRT}}

\section{Experiments}\label{sec:Experiments}

\subsection{Experimental setup}
% The citation to OpenHaRT should point to the summary report
% of OpenHaRT 2013?
Three handwriting datasets are used to evaluate our system:
% JL: I re-ordered the databases: from the smallest to the biggest % And yes, swapping columns in LaTeX tables was a pain in the ass ;-)
Rimes~\cite{Grosicki2009},
IAM~\cite{Marti2002IAM}
and OpenHaRT~\cite{OpenHaRT2013} containing handwritten
French,
English
and Arabic text, respectively.
%
%% JL: the following sentence is not clear
%They are multi-purpose datasets with several subsets dedicated for different recognition problems.
%
%% JL: do not deeply in the details, dataset at A2iA is a mess: we actually don't use any of the official splits...
%For OpenHaRT, 
%only a subset of the full training dataset was used for word recognition,
%but the full training dataset was used for line recognition. 
We split the databases into disjoint subsets to train, validate and evaluate our models.
The size of the selected datasets are given in Table~\ref{tab:Sizes-of-datasets}.
All the images used in these experiments consist of either isolated words (Section~\ref{ssec:expes_words})
or isolated lines (Section~\ref{ssec:expes_lines}).
They are all scanned at (or scaled to) 300\,dpi,
and we recall that the network architecture
presented in section~\ref{sec:RNN-intro} is designed to fit with this resolution.
%
%% JL: Could be interesting, but has to come fluidly in the discussion. -- Maybe out of scope here: Let's see if we display results on line...
%In fact, following the idea of curriculum learning \cite{bengio2009CurriculumLearning},
%we first train the model on isolated words and
%then continue training them on full line of text.
%
%% JL: not clear, because word->line needs adaptation (which is a modification!). -- Maybe out of scope here: Let's see if we display results on line...
%Thanks to the flexibility of the RNN architecture we are using,
%it does not require any modification to be used for full line recognition,
%which is also our ultimate goal.
%
\begin{table}[h]
\begin{threeparttable}
\caption{
The number of isolated words and lines in the datasets used in this work.
%The number of characters are also indicated in parenthesis
}
\label{tab:Sizes-of-datasets}
\centering
%\scriptsize
\begin{tabular}
%{|r||r||r||r|}
{|r||rr||rr||rlr|}
%\hhline{|-||-||-||-|}
\hhline{~|--||--||---|}
\multicolumn{1}{c|}{} & \multicolumn{2}{c||}{\Rimes} & \multicolumn{2}{c||}{\IAM} & \multicolumn{3}{c|}{\OpenHaRT}\\
\multicolumn{1}{c|}{} & \textbf{words} & \textbf{lines} & \textbf{words} & \textbf{lines} & \textbf{words} && \textbf{lines} \\
%subset & \Rimes & \IAM & \OpenHaRT \\ 
%\hhline{:=::=::=::=:}
\hhline{.-::==::==::===:}
\textbf{Training}
            &     44\,197 %(230k characters)%230\,259 characters
                  & 1\,400
            &     80\,421 %(338k characters)%338\,788 characters 
                  & 6\,482
            &    524\,196 & \hspace{-0.3cm}\tnote{1} %(2\,300k characters)%2\,299\,839 characters   %&  93\,450 lines
                  & \hspace{-0.15cm} 747\,676
\\
%             & (338k characters) & (230k characters) & (2\,300k characters) \\
\textbf{Validation}
            &  7\,542 %
                  & 100
            & 16\,770 
                  & 976
            & 57\,462 & %%& 9\,525
                  & 9\,525
\\
\textbf{Evaluation}
            &  7\,464 %
                  & 100
            & 17\,991
                  & 2\,915
            & 48\,308 & %%& 8\,483
                  & 8\,483
\\
%\hhline{|-||-||-||-|}
\hhline{|-||--||--||---|}
\end{tabular}
\begin{tablenotes}
\item [1] For OpenHaRT, only a subset of the full available data was used in the experiments on isolated word.
\end{tablenotes}
\end{threeparttable}
\end{table}
%

% Do we have any reference for CER and WER?
To assess the performance of our system, we measure %present
the Character Error Rate (CER) and Word Error Rate (WER).
The CER is computed by normalizing the total edit distance
between every pair of target and recognized sequences of characters (including the white spaces for line recognition). % given by the RNN.
The WER is simply the classification error rate in the case of isolated word recognition,
%where an error is counted if the recognized word is not exactly the same as the target word.
and is a normalized edit distance between sequences of words in the case of line recognition.

The RNN optical models are trained by online stochastic gradient descent with
a fixed learning rate of $10^{-3}$. The objective function is the
Negative Log-Likelihood (NLL) computed by CTC. All the weights
are initialized by sampling from a %normal distribution $\mathcal{N}\left(0,0.01\right)$
Gaussian distribution with zero mean and a standard deviation of $10^{-2}$. 
%a zero-mean Gaussian distribution with standard deviation 0.1.
A simple early stopping strategy is
employed and no other regularization methods than dropout were used.
When dropout
is enabled, we always use the dropout probability $p=0.5$.
%Thanks to a parallel implementation of gradient descent which executes simultaneously on 4 CPU threads, the network took several days to converge, depending on the size of the training set.

% A macro to condense the tables...
\def\condense{ \hspace{-0.3cm} }

\begin{table*}
\parbox[t]{0.45\textwidth}{
   \vspace{0pt}
   \begin{threeparttable}[b]
   \caption{Evaluation results of Word Recognition, with and without dropout at the topmost LSTM hidden layer}
   \label{tab:dropout-top-layer}
   \begin{tabular}{|cl||rr||rr||rr|}
   \hhline{--||--||--||--}
    \textbf{\# topmost} & \condense \textbf{Dropout}
      & \multicolumn{2}{c||}{\Rimes}
      & \multicolumn{2}{c||}{\IAM}
      & \multicolumn{2}{c|}{\OpenHaRT}
   \\
   \hhline{~~||~~||~~||~~}
    \textbf{LSTM cells} & \condense \textbf{on top}\condense & \CER & \WER & \CER & \WER & \CER & \WER \\
   \hhline{:==::==::==::==:}
   30  & No 
      & 14.72          & \textbf{42.03} 
      &\textbf{20.07} & \textbf{42.40} 
      & 12.80          & 37.44
      \\
   50  &    
      & 15.11          & 42.62          
      & 21.12         & 43.92          
      & 12.89          & 37.50
      \\
   100 &    
      & 15.79          & 44.37          
      & 21.87         & 43.82         
      & \textbf{12.48} & \textbf{36.50}
      \\
   200 &    
      & \textbf{14.68} & 42.07          
      & 22.23         & 44.83          
      & 13.14          & 37.65
      \\
   \hhline{:==::==::==::==:}
   30  & Yes 
      & 12.33          & 37.12          
      & 18.62          & 39.61          
      & 15.68          & 43.09
      \\
   50  & 
      & \textbf{12.17} & \textbf{36.03} 
      & \textbf{18.45} & 39.58          
      & 12.87          & 36.56
      \\
   100 & 
      & 12.20          & \textbf{36.03} 
      & 18.62          & \textbf{39.48} 
      & 11.50          & 33.71
      \\
   200 & 
      & 13.24          & 38.36          
      & 19.72          & 41.32          
      & \textbf{10.97} & \textbf{32.64}
      \\
   \hhline{--||--||--||--}
   \end{tabular}
   \begin{tablenotes}
   \item Bold numbers indicate the best results obtained for a given database and a given configuration.
   \end{tablenotes}
   \end{threeparttable}
}
\hfill
\parbox[t]{0.48\textwidth}{
   \vspace{0pt}
   \caption{Evaluation results of Word Recognition, with dropout at multiple layers}
   \label{tab:dropout-multi-layers}
   \begin{tabular}{|ll||rr||rr||rr|}
   \hhline{--||--||--||--|}
   \textbf{\# LSTM} & \condense \textbf{\# LSTM layers}
      & \multicolumn{2}{c||}{\Rimes} 
      & \multicolumn{2}{c||}{\IAM} 
      & \multicolumn{2}{c|}{\OpenHaRT}
      \\
   \hhline{~~||~~||~~||~~|}
   \multicolumn{1}{|c}{\textbf{cells}} & \textbf{with dropout} & \CER & \WER & \CER & \WER & \CER & \WER \\
   \hhline{==::=:=::=:=::=:=:}

    2, 10, 50  &                       
      & 15.11 & 42.62 
      & 21.12 & 43.92          
      & 12.89 & 37.50
      \\
    2, 10, 100 &                       
      & 15.79 & 44.37 
      & 21.87 & 43.82          
      & 12.48 & 36.50\\
    2, 20, 50  & 0                     
      & \textbf{13.49} & \textbf{39.42} 
      & 20.67 & 42.20          
      & 11.32 & 33.96\\
    2, 20, 100 & {\it(no dropout)}     
      & \textbf{13.64} & \textbf{39.86} 
      & 19.79          & \textbf{41.22} 
      & 11.15 & 33.55 
      \\
    4, 20, 50  &                       
      & 14.48 & 41.65
      & \textbf{19.67} & \textbf{41.15} 
      & \textbf{10.93} & \textbf{32.84}
      \\
    4, 20, 100 &                       
      & 14.83  & 42.28 
      & \textbf{19.46} & 41.47 
      & 11.07 & 33.09
      \\
   \hhline{==::=:=::=:=::=:=:}
    2, 10, 50  & 1 {\it (topmost)}     
      & 12.17 & 36.03 
      & 18.45 & 39.58 
      & 12.87 & 36.56
      \\
    2, 10, 100 &                       
      & 12.20 & 36.03
      & 18.62 & 39.48
      & 11.50 & 33.71 \\
   \hhline{--||--||--||--}
    2, 20, 50  & 2 {\it (top)}        
      & \textbf{8.95}   & \textbf{28.70}
      & 14.52 & 32.32
      & 10.48 & 31.45
      \\
    2, 20, 100 &                       
      & 9.29  & 28.98 
      & 15.06 & 32.96 
      & \textbf{9.17} & \textbf{28.17} \\
   \hhline{--||--||--||--}
    4, 20, 50  & 3                     
      & \textbf{8.62}  & \textbf{27.01} 
      & \textbf{13.92} & \textbf{31.48} 
      & 11.21 & 33.11 \\
    4, 20, 100 &                       
      & 9.98  & 30.63 
      & \textbf{14.02} & \textbf{31.44} 
      & 9.77 & 29.58 \\
   \hhline{--||-|-||-|-||-|-|}
   \end{tabular}
}
\end{table*}

\subsection{Isolated Word Recognition}
\label{ssec:expes_words}

\subsubsection{Dropout at the topmost LSTM layer}% for word recognition}

In this set of experiments, we first apply dropout at the topmost LSTM layer.
Since there are 50 features at this layer, dropout can sample from a great
number of networks. Moreover,
since the inputs of this layer have smaller sizes than those of lower layers due to subsampling,
dropout at this layer will not take too much
time during training.

Previous work \cite{Hinton2012DropoutTalk} suggests that dropout
is most helpful when the size of the model is relatively big, and
the network suffers from overfitting. One way to control the size of
the network is to change the number of hidden features in the recurrent layers.
While the baseline architecture has 50 features at the topmost layer, we vary it
among 30, 50, 100 and 200. All other parameters are kept fixed,
the network is then trained with and without dropout.

For each setting and dataset, the model with highest performance on validation
set is selected and evaluated on corresponding test set. The results are given in
Table \ref{tab:dropout-top-layer}.
It can be seen that dropout works very well on IAM and Rimes where
it significantly improves the performance by $10-20\%$ % on a relative basis
regardless of the number of topmost hidden units. On OpenHaRT, dropout also helps
with 50, 100 or 200 units, but hurts the performance with 30 units, 
% This can be explained by the fact that the model with 30 units seems to be underfitted.
most likely because the model with 30 units is underfitted.

Fig.~\ref{fig:Convergence-curves-of-dropout} depicts the convergence
curves of various RNN architectures trained on the three datasets
when dropout is disabled or enabled.
In all experiments, convergence curves show that
dropout is very effective in preventing %the network from
overfitting. When dropout is disabled, the RNNs clearly
suffer from overfitting as their NLL on the validation
dataset increases after a certain number of iterations. When dropout
is enabled, the networks are better regularized
and can achieve higher performance on validation set at the end.
Especially for OpenHaRT,
since its training and validation sets are much larger
than IAM and Rimes, 30 hidden units are inadequate and training
takes a long time to converge. With 200 units and no dropout,
it seems to be overfitted. However when dropout is enabled, 200 units
give very good performance.

\subsubsection{Dropout at multiple layers}% for word recognition}

%In previous experiments, we only applied dropout at the topmost LSTM layer.
%In this section, we would like to explore
Now we explore the possibilities of using dropout
also at other layers than the topmost LSTM layer.
In our architecture, there are 3 LSTM layers, hence we tried
applying dropout at the topmost, the top two and all the three LSTM layers.

Normally when dropout is applied at any layer, we double the number of LSTM units
at that layer. This is
%% JL: I tried to clarify
%because the number of LSTM features in the standard architecture (2, 10 and 50) are well-tuned for the network without dropout.
%When dropout is applied, we need a bigger network, hence doubling the size of LSTM layers is reasonable.
to keep the same number of active hidden units (on average) when using dropout with~$p=0.5$
as in the baseline where all hidden units are active. %(and for which the number of units is supposed to be tuned for RNN without dropout).
We remind that the baseline architecture consists of LSTM layers with 2, 10 and 50 units,
so it would correspond to an architecture of 4, 20 and 100 units when dropout is applied at every layer.
Since most of free parameters of the networks concentrate at the top layers, 
doubling the last LSTM layer almost doubles the number of free parameters.
Therefore we also have several experiments where we keep the last LSTM layer at 50 units with dropout.
Besides, in order to avoid favouring the models trained with dropout because they have greater capacity,
we also test those big architectures without dropout.
%All the architectures are trained when dropout is disabled or applied
%at selective layers.

Their performance are reported in Table~\ref{tab:dropout-multi-layers}.
Since we double the size of LSTM layers, the modeling power of the RNNs is increased. Without dropout,
the RNNs with more features at lower layers generally obtain higher performance. However 
we observed overfitting on Rimes when we use 4 and 20 features at the lowest LSTM layers.
This makes sense because Rimes is the smallest of the three datasets. With dropout, CER and WER
decrease by almost 30-40\% on a relative basis. We found that dropout
at 3 LSTM layers is generally helpful, however the training time is significantly longer
both in term of the number of epochs before convergence and the CPU time for each epoch.

% TODO: need a plot for training time?

\subsection{Line Recognition with Lexical Constraints and Language Modeling}
\label{ssec:expes_lines}

Note that the results presented in Table \ref{tab:dropout-multi-layers} can not be directly compared 
to state-of-the-art results previously published on the same databases \cite{Kozielski2013a,A2iASystem_OpenHart2013},
%since only the output of the optical model is considered, \textit{i.e.} unconstrained sequences of characters. 
since the RNNs only output unconstrained sequences of characters. 
A complete system for large vocabulary handwriting text recognition includes a lexicon and a language model, 
which greatly decrease the error rate by inducing 
lexical constraints and rescoring the hypotheses produced by the optical model.

In order to compare our approach to existing results, we trained again the best
RNNs for each database, with and without dropout, on lines of text. 
The whitespaces in the annotations are also considered as targets for training.

Concretely, we build a hybrid HMM/RNN model. 
There is a one-state HMM for each label (character, whitespace, and the blank
symbol of CTC~\cite{Graves06connectionisttemporal}), which
has a transition to itself and an outgoing transition with the same probability.
The emission probabilities are obtained by transforming the posterior probabilities given by the RNNs
into pseudo-likelihood. Specifically, the posteriors $p(s|x)$ are divided by the priors $p(s)$,
scaled by some factor $\kappa$ : $\dfrac{p(s|x)}{p(s)^\kappa}$, where $s$ is the
HMM state, i.e. a character, a blank, or a whitespace, and $x$ is the input.
The priors $p(s)$ are estimated on the training set.

We include the lexical contraints (vocabulary and language model)
in the decoding phase as a Finite-State Transducer (FST),
which is the decoding graph in which we inject the RNN predictions. The method to create
an FST that is compatible with the RNN outputs is described in~\cite{A2iASystem_OpenHart2013}.
The whitespaces are treated as an optional word separator in the lexicon.
The HMM is also represented as an FST $H$ and is composed with the lexicon FST $L$,
and the language model $G$. 

The final graph $HLG$ is the decoding graph in which we search the best sequence of
words $\mathbf{\hat{W}}$
\begin{equation*}
 \mathbf{\hat{W}} = arg\max_\mathbf{W} [\omega \log p(\mathbf{X}|\mathbf{W}) + \log p(\mathbf{W}) + |\mathbf{W}| \log WIP ]
\end{equation*}
where $\mathbf{X}$ is the image,
$p(\mathbf{X}|\mathbf{W})$ are the pseudo-likelihoods, 
$p(\mathbf{W})$ is given by the language model,
$\omega$ and $WIP$ are the optical scaling factor -- balancing the importance
of the optical model and the language model -- and the word insertion penalty.
These parameters, along with the prior scaling factor $\kappa$, have been tuned 
independently for each database on its validation set.

For IAM, we applied a $3$-gram language model
trained on the LOB, Brown and Wellington corpora. 
The passages of the LOB corpus appearing in the 
validation and evaluation sets were removed prior to LM training. 
We limited the vocabulary to the 50k most frequent words.
The resulting model has a perplexity of 298 and OOV rate of 4.3\% on the validation set 
(329 and 3.7\% on the evaluation set).

For Rimes, we used a vocabulary made of 12k words from the training set.
We built a $4$-gram language model with modified Kneser-Ney discounting from the training annotations.
The language model has a perplexity of 18 and OOV rate of 2.6\% on the evaluation set.

For OpenHaRT, we selected a 95k words vocabulary containing all the words of the 
training set. 
We trained a $3$-gram language model on the training set annotations,
with interpolated Kneser-Ney smoothing. 
% The vocabulary selection is described with more details in~\cite{A2iASystem_OpenHart2013}.
The language model has a perplexity of 1162 and OOV rate of 6.8\% on the evaluation set.

\begin{table*}
\parbox[t]{0.33\textwidth}{
    %\begin{table}[!htb]
    \vspace{0pt}
    \caption{Results on Rimes}
    \label{tab:finalresultsrimes}
    \centering
    \begin{tabular}{|r||c|c||c|c|}
    \hhline{~|-|-||-|-|}
    \multicolumn{1}{r|}{} & \multicolumn{2}{c||}{\textbf{Valid.}} & \multicolumn{2}{c|}{\textbf{Eval.}} \\
    \hhline{~|~~||~~|}
    \multicolumn{1}{r|}{}  & \condense \WER \condense & \condense \CER \condense & \condense \WER \condense & \condense \CER \condense \\
    \hhline{-::=:=::=:=:}
    \multicolumn{1}{|l||}{MDLSTM-RNN}  & 32.6 & 8.1 & 35.4 & 8.9 \\
    \multicolumn{1}{|l||}{~~+ dropout} & 25.4 & 5.9 & 28.5 & 6.8 \\
    \multicolumn{1}{|l||}{~~+ Vocab\&LM} & 14.0 & 3.7 & 12.6 & 3.5 \\
    \multicolumn{1}{|l||}{~~~~+ dropout} & \textbf{13.1} & \textbf{3.3} & \textbf{12.3} & \textbf{3.3} \\
    \hhline{=::=:=::=:=:}
    % Bluche et al.~\cite{BlucheICPR}   & \textbf{11.8} & \textbf{3.7} \\
    % Bluche et al.~\cite{BlucheICPR}   & 12.7 & 4.0 \\
    Messina et al.~\cite{MessinaOOV2014} & - & - & 13.3 & - \\
    Kozielski et al.~\cite{Kozielski2013a} & - & - &  13.7 & 4.6 \\
    Messina et al.~\cite{MessinaOOV2014} & - & - &  14.6 & - \\
    Menasri et al.~\cite{Menasi2012}  & - & - &  15.2 & 7.2 \\
    % & Telecom ParisTech.~\cite{Grosicki2011} & - & - & 31.2 & 18.0 \\
    \hhline{-||-|-||-|-|}
    \end{tabular}
    %\end{table} 
}
\enskip
\parbox[t]{0.33\textwidth}{
    %\begin{table}[!htb]
    \vspace{0pt}
    \caption{Results on IAM}
    \label{tab:finalresultsiam}
    \centering
    \begin{tabular}{|r||c|c||c|c|}
    \hhline{~|-|-||-|-|}
    \multicolumn{1}{r|}{} & \multicolumn{2}{c||}{\textbf{Valid.}} & \multicolumn{2}{c|}{\textbf{Eval.}} \\
    \hhline{~|~~||~~|}
    \multicolumn{1}{r|}{}  & \condense \WER \condense & \condense \CER \condense & \condense \WER \condense & \condense \CER \condense \\
    \hhline{-::=:=::=:=:}
    \multicolumn{1}{|l||}{MDLSTM-RNN}  & 36.5 & 10.4 & 43.9 & 14.4 \\
    \multicolumn{1}{|l||}{~~+ dropout} & 27.3 & 7.4 & 35.1 & 10.8 \\
    \multicolumn{1}{|l||}{~~+ Vocab\&LM} & 12.1 & 4.2 & 15.9 & 6.3 \\
    \multicolumn{1}{|l||}{~~~~+ dropout} & 11.2 & 3.7 & 13.6 & \textbf{5.1} \\
    \hhline{=::=:=::=:=:}
    % Bluche et al.~\cite{BlucheICPR}    & 9.7  & 3.6 & \textbf{11.9} & \textbf{4.9} \\
    Kozielski et al.~\cite{Kozielski2013a} & \textbf{9.5} & \textbf{2.7} & \textbf{13.3} & \textbf{5.1} \\
    Kozielski et al.~\cite{Kozielski2013a} & 11.9 & 3.2 & - & - \\
    % Bluche et al.~\cite{BlucheICPR}    & 11.9  & 3.9 & 14.3 & 5.3 \\
    Espana et al.~\cite{Espana-Boquera_etal2010} & 19.0 & - & 22.4 & 9.8 \\
    Graves et al.~\cite{Graves2009a} & - & - & 25.9 & 18.2 \\
    Bertolami et al.~\cite{Bertolami_Bunke2008} & 26.8 & - & 32.8 & - \\
    Dreuw et al.~\cite{Dreuw2011} & 22.7 & 7.7 & 32.9 & 12.4 \\
    \hhline{-||-|-||-|-|}
    \end{tabular}
    %\end{table} 
}
\enskip
\parbox[t]{0.33\textwidth}{
    %\begin{table}[!htb]
    \vspace{0pt}
    \caption{Results on OpenHaRT}
    \label{tab:finalresultsopenhart}
    \centering
    \begin{tabular}{|r||c|c||c|c|}
    \hhline{~|-|-||-|-|}
    \multicolumn{1}{r|}{} & \multicolumn{2}{c||}{\textbf{Valid.}} & \multicolumn{2}{c|}{\textbf{Eval.}} \\
    \hhline{~|~~||~~|}
    \multicolumn{1}{r|}{}  & \condense \WER \condense & \condense \CER \condense & \condense \WER \condense & \condense \CER \condense \\
    \hhline{-::=:=::=:=:}
    \multicolumn{1}{|l||}{* MDLSTM-RNN}  & 31.0 & 7.2 & 34.7 & 8.4 \\
    \multicolumn{1}{|l||}{*~+ dropout} & 27.8 & 6.4 & 30.3 & 7.3  \\
    \multicolumn{1}{|l||}{~~+ Vocab\&LM} & 8.3 & 3.8 & 18.6 & 4.9  \\
    \multicolumn{1}{|l||}{~~~~+ dropout} & 8.2 & 3.8 & \textbf{18.0} & \textbf{4.7}  \\
    \hhline{=::=:=::=:=:}
    Bluche et al.~\cite{A2iASystem_OpenHart2013}    & - & - & 23.3 & - \\
    Bluche et al.~\cite{A2iASystem_OpenHart2013}    & - & - & 25.0 & - \\
    Kozielski et al.~\cite{hamdani14openhart}       & - & - & 25.8 & 10.7 \\
    \hhline{-||-|-||-|-|}
    \end{tabular}
    %\end{table} 
    \begin{tablenotes}
    \item * The error rates in the first 2 lines are computed \\
    from the decomposition into presentation forms \\
    and are not directly comparable to the remaining of \\
    the table.
    \end{tablenotes}
}
\end{table*}

\def\Baseline{\textbf{Baseline}}
\def\Dropout{\textbf{Dropout}}

\begin{table*}[t!]
\centering
\parbox[t]{0.45\textwidth}{
   \vspace{0pt}
   \begin{threeparttable}
   \caption{Norm of the weights, for differently trained RNNs.}
   \label{tab:Norm-of-weights}
   \begin{tabular}{|ll||rr||cc||ll|}
   \hhline{~~|--||--||--|}
   \multicolumn{2}{c|}{} 
      & \multicolumn{2}{c||}{\Rimes} 
      & \multicolumn{2}{c||}{\IAM} 
      & \multicolumn{2}{c|}{\OpenHaRT}\tabularnewline
   %\cline{2-7} 
   \hhline{~~|~~||~~||~~|}
   \multicolumn{2}{c|}{} & \condense \Baseline \condense & \condense \Dropout \condense & \condense \Baseline \condense & \condense \Dropout \condense & \condense \Baseline \condense & \condense \Dropout\condense \tabularnewline
   \hhline{--::==::==::==:}
   \textbf{LSTM} & \condense \textbf{L1-norm}
      & 0.162 & 0.213 
      & 0.181 & 0.220
      & 0.259 & 0.307 \\
   \multicolumn{1}{|r}{\textbf{weights}} & \condense \textbf{L2-norm}
      & 0.200 & 0.263
      & 0.225 & 0.273 
      & 0.322 & 0.382 \\
   \hhline{:==::==::==::==:}
   \textbf{Classif.} & \condense \textbf{L1-norm}
      & 0.152 & 0.097
      & 0.188 & 0.113
      & 0.277 & 0.175 \\
   \multicolumn{1}{|r}{\textbf{weights}} & \condense \textbf{L2-norm}
      & 0.193 & 0.120
      & 0.238 & 0.139
      & 0.353 & 0.215 \\
   \hhline{|--||--||--||--|}
   \end{tabular}
   \begin{tablenotes}
   \item The first 2 lines correspond to weights in the topmost LSTM layer (before dropout, if any)
   and the last 2 lines correspond to classification weights in topmost linear layer (after dropout, if any). %, before Softmax non-linear function.
   \end{tablenotes}
   \end{threeparttable}
}
\hfill
\parbox[t]{0.5\textwidth}{
    \vspace{0pt}
    \includegraphics[width=0.5\textwidth]{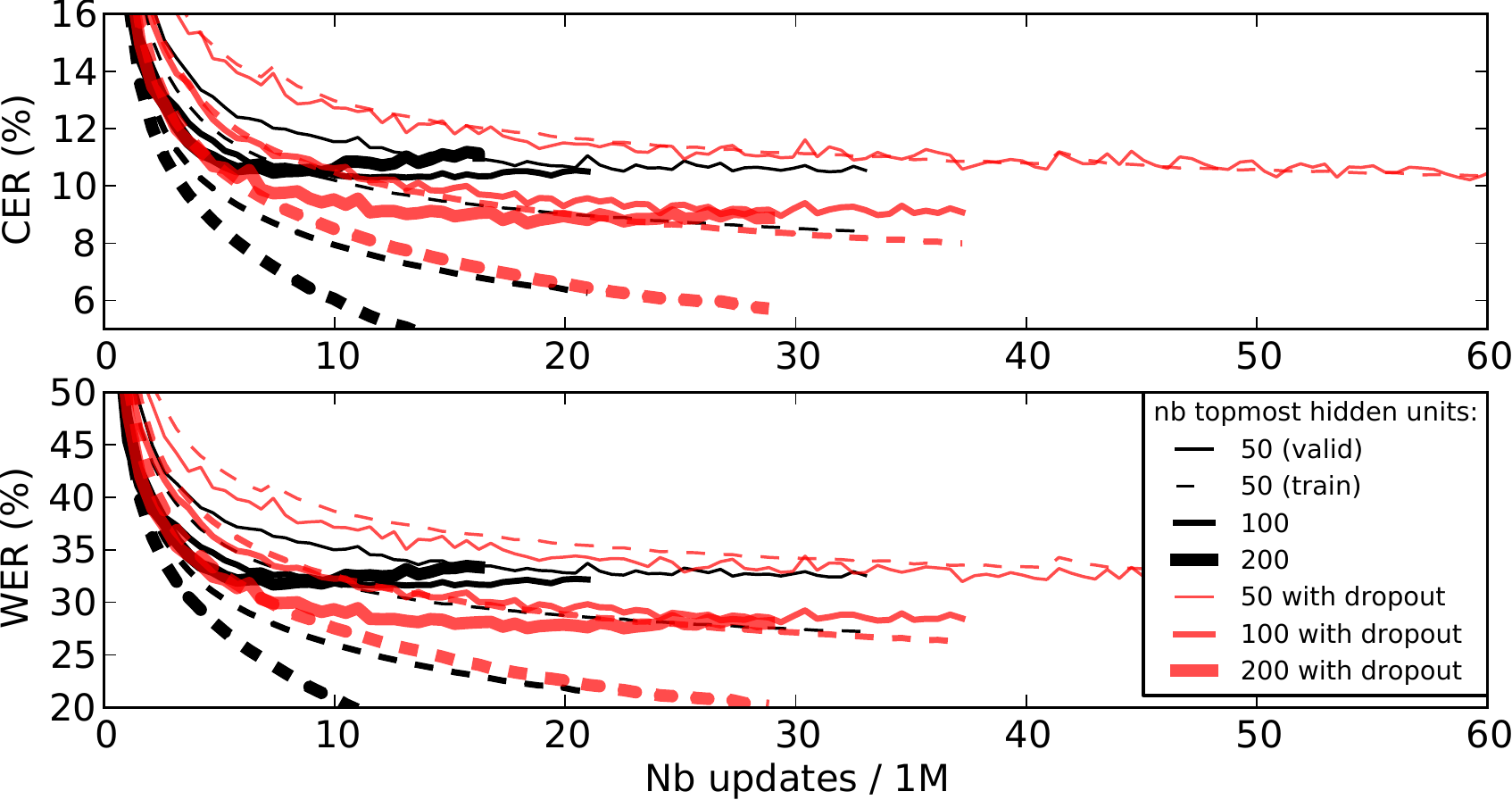}
    \captionof{figure}{
    Convergence Curves on OpenHaRT.
    Plain ({\it resp.} dashed) curves show the costs on the validation ({\it resp.} training) dataset.
    }
    \label{fig:Convergence-curves-of-dropout}
}
\end{table*}

The results are presented in Tables~\ref{tab:finalresultsrimes} (Rimes), 
\ref{tab:finalresultsiam}~(IAM) and \ref{tab:finalresultsopenhart}~(OpenHaRT).
On the first two rows, we present the error rates of the RNNs alone, without any
lexical constraint. It can be seen that dropout gives from 7 to 27\% relative improvement.
The third rows present the error rates when adding lexical constraints without dropout.
In this case, only valid sequences of characters are outputed,
and the relative improvement in CER over the
systems without lexical constraints is more than 40\%.
On the 4th row, when dropout and lexical constraints are both enabled, dropout achieves
5.7\% (Rimes), 19.0\% (IAM) and 4.1\% (OpenHaRT) relative improvement in CER, and 
2.4\% (Rimes), 14.5\% (IAM) and 3.2\% (OpenHaRT) relative improvement in WER.
Using a single model and closed vocabulary, our systems outperform
the best published results for all databases.
Note that on the 5th line of Table~\ref{tab:finalresultsiam},
the system presented in~\cite{Kozielski2013a}
adopts an open-vocabulary approach and can recognize out-of-vocabulary words,
which can not be directly compared to our models.

\subsection{Effects of dropout on the Recurrent Neural Networks}

In order to better understand the behaviour of dropout in training RNNs,
we analyzed the distribution of the network weights and the intermediate activations.
Table~\ref{tab:Norm-of-weights} shows the L1 and L2 norm of the weights
of LSTM gates and cells in the topmost LSTM layer (referred to as "LSTM weights"),
and the weights between the topmost LSTM layer and the softmax layer ("Classification weights").
It is noticeable that the classification weights are smaller when dropout
is enabled. We did not use any other regularization method, but
\emph{dropout seems to have similar regularization effects as L1 or L2 weight decay}.
The nice difference is that the hyper-parameter~$p$ of dropout is much less tricky to tune than those of weight decay.

On the other hand,
the LSTM weights tend to be higher with dropout,
and further analysis of the intermediate activations shows that
%when dropout is enabled, %the distribution of activations has a more rounded peak and greater standard deviation, indicating that 
the distribution of LSTM activations have a wider spread.
This side effect can be partly explained by the hypothesis that dropout \emph{encourages the units to emit
stronger activations}. Since some units were randomly dropped during training,
stronger activations might make the units more independently
helpful, given the complex contexts of other hidden activations.
Furthermore, we checked that the LSTM activations are not saturated under the effect of dropout.
Keeping unsaturated activations is particularly important 
%when training LSTM cells because it allows error gradient
%to be passed during the backward pass of gradient descent, which in turn
%enables the recurrent connections to learn long-term dependencies in the training sequences.
when training RNN, since it ensures that the error gradient can be propagated to learn long-term dependencies.

The regularization effect of dropout is certain when we look into the
learning curves given in Fig.~\ref{fig:Convergence-curves-of-dropout}, where it shows how overfitting
can be greatly reduced.
%Besides, the less there is data in the training set, the more dropout can improve performance relatively.
The gain of dropout becomes highly significant when the network gets relatively bigger with respect to the dataset.

% Due to space constraints, we can not detail the results of such a system  but  
%  we obtained an improvements of 15-20\% in CER when dropout was used  in a    
%  large vocabulary Arabic handwritten text recognition system~\cite{A2iASystem_OpenHart2013}.
%Finally, using a similar network architecture for recognizing lines of text (there are more than
%one word in each image snippet), we obtained an improvements of 15-20\% in CER when dropout is enabled
%\footnote{We cannot describe those results in detail due to space constraints, but the full system
%is presented in~\cite{A2iASystem_OpenHart2013}.}.

\section{Conclusion}

%% JL: We apply dropout, not an "enhanced" version, don't we?
%We presented an enhanced version of dropout which works nicely with both recurrent in deep architectures.
We presented how dropout can work with both recurrent and convolutional layers in a deep network architecture.
The word recognition networks with dropout at the topmost layer
significantly reduces the CER and WER by 10-20\%, and the performance can be further improved
by 30-40\% if dropout is applied at multiple LSTM layers.
The experiments on complete line recognition also showed that dropout always improved the
error rates, whether the RNNs were used in isolation, or constrained by a lexicon and a
language model. We report the best known results on Rimes and OpenHaRT databases.
Extensive experiments also provide evidence that dropout behaves 
similarly to weight decay, but the dropout hyper-parameter is much easier to tune than those of weight decay.
It should be noted that although our experiments were conducted on handwritten datasets,
the described technique is not limited to handwriting recognition,
it can be applied as well in any application of RNNs. %deep neural networks, especially RNNs.

\section*{Acknowledgement}
This work was partially funded by the French Grand Emprunt-Investissements d'Avenir program through the PACTE project,
and was partly achieved as part of the Quaero Program, funded by OSEO, 
French State agency for innovation.


\bibliographystyle{IEEEtran}
\bibliography{IEEEabrv,biblio}

\end{document}